\documentclass[9pt]{article}
\usepackage{amsmath,graphicx}
\usepackage[preprint]{spconf}
\usepackage{stfloats,framed}
\usepackage{hyperref}

\usepackage{caption}
\usepackage{subcaption}

\title {{A Robust Quantile Huber Loss with Interpretable Parameter Adjustment in Distributional Reinforcement Learning 
 }}

\name{ Parvin Malekzadeh~$^{\star \dagger}$ \thanks{$^{\star}$ Corresponding author} \quad Konstantinos N. Plataniotis~$^{\dagger}$ \quad  Zissis Poulos~$^{\ddagger}$ \quad Zeyu Wang~$^{\ddagger}$ \qquad \qquad \qquad \qquad \qquad \qquad \qquad \qquad}
\address{{\small  $^{\dagger}$ The Edward S. Rogers Sr. Department of Electrical and Computer Engineering, University of Toronto, ON, Canada  \qquad\qquad\qquad\qquad \qquad \qquad\qquad \qquad \qquad \qquad \qquad} \\
			   {\small $^{\ddagger}$ Joseph L. Rotman School of Management, University of Toronto, Toronto, ON, Canada \qquad \qquad\qquad \qquad \qquad\qquad\qquad \qquad \qquad \qquad\qquad\qquad }}
\usepackage{microtype}
\usepackage{amsmath}
\usepackage{multirow}
\usepackage{amsthm}
\usepackage{amssymb }
\usepackage{multirow}
\usepackage{amsmath,tabularx}
\usepackage{bm}
\usepackage[numbers]{natbib}
\usepackage{float}
\usepackage{amsfonts}
\usepackage{flushend}
\usepackage{geometry}
\usepackage{mathtools}
\usepackage[dvipsnames]{xcolor}
\usepackage{algorithm, algpseudocode}

\newcommand{\overbar}[1]{\mkern 1.mu\overline{\mkern-1.mu#1\mkern-1.mu}\mkern 1.mu}
\def\mS{\mathcal{S}}
\def\mA{\mathcal{A}}

\def\mS{\mathcal{S}}

\def\mA{\mathcal{A}}

\def\bt{{\theta}}

\def\s{\bm{s}}
\def\pk{_{k-1}}
\def\pt{_{t-1}}

\def\t{_t}

\newcommand\myeq{\stackrel{\mathclap{\normalfont\mbox{D}}}{=}}
\makeatletter
\newcommand{\multiline}[1]{%
  \begin{tabularx}{\dimexpr\linewidth-\ALG@thistlm}[t]{@{}X@{}}
    #1
  \end{tabularx}
}
\begin{document}
%
\maketitle
\begin{figure*}[b!]
\begin{framed}
Copyright 2024 IEEE. To appear in ICASSP 2024 - 2024 IEEE International Conference on Acoustics, Speech and Signal Processing (ICASSP), scheduled for 14-19 April 2024 in Seoul, Korea. Personal use of this material is permitted. However, permission to reprint/republish this material for advertising or promotional purposes or for creating new collective works for resale or redistribution to servers or lists, or to reuse any copyrighted component of this work in other works, must be obtained from the IEEE. Contact: Manager, Copyrights and Permissions / IEEE Service Center / 445 Hoes Lane / P.O. Box 1331 / Piscataway, NJ 08855-1331, USA. Telephone: + Intl. 908-562-3966.
\end{framed}
\end{figure*}
\begin{abstract}
\vspace{-.03in} Distributional Reinforcement Learning (RL) estimates return distribution mainly by learning quantile values via minimizing the quantile Huber loss function, entailing a threshold parameter often selected heuristically or via hyperparameter search, which may not generalize well and can be suboptimal.
This paper introduces a generalized quantile Huber loss function derived from Wasserstein distance (WD) calculation between Gaussian distributions, capturing noise in predicted (current) and target (Bellman-updated) quantile values. Compared to the classical quantile Huber loss, this innovative loss function enhances robustness against outliers.  Notably, the classical Huber loss function can be seen as an approximation of our proposed loss, enabling parameter adjustment by approximating the amount of noise in the data during the learning process.
Empirical tests on Atari games, a common application in distributional RL, and a recent hedging strategy using distributional RL, validate the effectiveness of our proposed loss function and its potential for parameter adjustments in distributional RL. The implementation of the proposed loss function is available \href{https://github.com/pmalekzadeh/A-robust-quantile-huber-loss}{here}. 
\end{abstract}
%
\vspace{-.19in} \section{Introduction}
\label{sec:intro}\vspace{-.16in}
Within the domain of reinforcement learning (RL)~\cite{malekzadeh2023uncertainty}, distributional RL~\cite{bellemare2017distributional, dabney2018distributional} revolves around estimating the distribution of (discounted) cumulative future rewards, known as the return. By explicitly delving into the distribution of returns and accounting for sensitivity to risk~\cite{ clavier2022robust, malekzadeh2023unified}, distributional RL has garnered interest in shaping optimal hedging strategies~\cite{cao2023gamma} to mitigate risk within dynamic and uncertain market landscapes. 
Any distributional RL algorithm involves two key elements: modelling the return distribution and selecting an optimization loss function~\cite{dabney2018implicit}.
 C51~\cite{bellemare2017distributional} and D4PG~\cite{barth2018distributed} use categorical return distributions with the projected Kullback-Leibler (KL) divergence between target (Bellman-updated) and current return distributions as their loss function. 
QR-DQN~\cite{dabney2018distributional} and IQN~\cite{dabney2018implicit} learn quantile representation of the return distribution by approximating quantile values via the 1-Wasserstein distance (WD), employing the quantile Huber loss~\cite{huber1992robust} (also known as asymmetric Huber loss). Unlike C51, QR-DQN and IQN achieve substantial performance gains without projection.
The quantile Huber loss blends robustness against outliers from the Huber loss with the ability to capture heterogeneity in returns through asymmetric quantile loss~\cite{zhao2021robust}. Modelling the heterogeneity is crucial when different parts of the distribution have distinct implications, such as in financial tasks~\cite{zhao2021robust}. This loss function employs a threshold parameter to control computations across distribution segments.
Expanding on the quantile Huber loss, FQF~\cite{yang2019fully} learns both quantile fractions and values, outperforming QR-DQN and IQN in Atari Games. Moreover, D4PG-QR~\cite{cao2023gamma} improves upon D4PG by using quantile Huber loss instead of KL divergence. These algorithms use a fixed threshold of $1$ in their quantile Huber loss, which can be suboptimal and may limit generalizability. We aim to offer a data-driven interpretation for this shared parameter, ensuring adaptability across diverse tasks.
\\
\textbf{Contributions:}
We provide a probabilistic interpretation of the quantile Huber loss, casting it as an asymmetric version of the 1-WD between Gaussian distributions, representing noise in predicted and target quantiles. This insight leads to a generalized quantile Huber loss with the following attributes: \\
1. The proposed loss function encompasses the classical quantile Huber loss as an approximation and offers enhanced robustness to outliers and smoother differentiability, resulting in faster convergence. \\
2. The embedded parameter in the proposed loss function is intrinsically linked to uncertainties in the predicted and target quantiles, allowing adaptive tuning for specific problem characteristics. \\
3.  In Atari games and option hedging experiments, our loss function, with its robustness and parameter adaptability, outperforms algorithms using the quantile Huber loss with the fixed threshold of $1$. Moreover, while the performance of these algorithms varies with the threshold parameter's deviation from $1$, requiring grid search, our loss function efficiently identifies the optimal parameter value.
%
\\
\textbf{Related Work:}
Prior research introduced robust, generalized forms of the quantile Huber loss, widely applied in simulated regression tasks~\cite{zhao2021robust, gokcesu2021generalized, taggart2022point, taggart2022evaluation, tyralis2023deep}. Unlike our single-parameter approach, these variants typically require two thresholds and involve complex two-dimensional grid searches due to their lack of intuitive parameter interpretations.
   Patterson et al.~\cite{patterson2022robust} introduced a reformulation of the Huber loss in RL but did not integrate it with quantile loss for quantile learning or propose parameter adjustments.
In related works~\cite{noy2014robust, meyer2021alternative}, authors noted the similarity between the Huber loss and the KL divergence of Laplace distributions in regression prediction. They fine-tuned the Huber loss threshold parameter based on this connection.  In contrast, our work focuses on the WD as the loss function, which converges to zero during quantile learning, and its correlation with the quantile Huber loss in distributional RL. Moreover, while they used this connection for parameter selection, our contribution goes further by introducing a generalized quantile Huber loss function, proven to be smoother and more robust to outliers than the conventional quantile Huber loss.
Alternative methods~\cite{choi2019distributional, nam2021gmac} use Gaussian distributions for return distribution and Gaussian-specific loss functions. However, we model the return distribution using quantiles and assume a Gaussian noise in these quantiles.  While Gaussian modeling offers simplicity and smoothness, quantile modeling provides robustness against outliers and insights into quantiles.  Hence, ongoing work in distributional RL~\cite{clavier2022robust,cao2023gamma,  oh2023toward, he2022popo} continues using quantile modeling. 
  \vspace{-.35in}\section{Distributional RL} \label{sec:background}
 \vspace{-.15in}
 Following common methods in RL, we model an agent-environment interaction with a Markov Decision Process (MDP) specified by $(\mS, \mA, P, \gamma, r) $, where $\mS$ and $\mA$ are state and action spaces, $P(.|\s,a)$ is the transition distribution, $\gamma \!\!\! \in \!\!\!(0,1]$ is the discount factor, and $r(\s,a)$ is the reward function. A policy $\pi(.|s)$ maps a state to a distribution over $\mA$. The objective in RL to find an optimal policy $\pi^*$ maximizing the expected return $ Q^{\pi}\!(\s, a) \!\!=\!\! \mathbb{E} [\sum_{t=0}^{\infty} \! \gamma^{t} r(\s\t,a\t)|s_0 \!= \!s,a_0 \!= \!a]$ for all $s,a$, where $\s\t \!\! \sim \!\! P(.| \s\pt, a\pk)$ and $a\t \! \! \sim \! \pi(.|\s\t)$.
In distributional RL, the goal is to learn the distribution of the return $Z^{\pi}(\s,a) \!\! = \! \!\sum_{t=0}^{\infty} \! \gamma^{t} r(\s\t,a\t)$. The return distribution for a policy $\pi$ can be computed using the distributional Bellman equation~\cite{bellemare2017distributional} as:
\vspace{-.1in}
{\small
\begin{equation}
Z^{\pi}(\s, a) \myeq T^{\pi} Z(\s, a) :\myeq  r(\s,a)+ \gamma Z^{\pi}(S', A'), \label{Eq:Bellman}
\end{equation}} \normalsize \par  \vspace{-.07in}
\noindent{where}  $T^{\pi}$ denotes the distributional Bellman operator, $ {S' \!\!   \sim \!\! P(.| \s, a)} $, and $ A' \! \! \sim \! \pi(.|S')$.
\vspace{.01in} \\
\textbf{Quantile Huber Loss for Distributional RL:}
Bellemare et al.~\cite{bellemare2017distributional} showed that the distributional Bellman operator $T^{\pi}$ in Eq.~\eqref{Eq:Bellman} is a contraction in the $p$-WD, i.e., applying $T^{\pi}$ iteratively, starting from an initial distribution of $Z$, convergences to the true distribution of $Z^{\pi}$ under the $p$-WD. However, they noted that minimizing the WD using samples introduces bias, making WD minimization impractical with gradient methods. Later, Dabney et al.~\cite{dabney2018distributional} approximated the distribution of $Z^{\pi}\!(s, a)$ as a mixture of $N$ Diracs located at $\bt_{\bm{\psi}}^{(1)}\!(s,a), \bt_{\bm{\psi}}^{(2)}\!(s,a), ..., \bt_{\bm{\psi}}^{(N)}\!(s,a)$ with parameters $\bm{\psi}$:
\vspace{-.18in}{
\begin{equation}
\!\! Z_{\bm{\psi}}(\s, a) := \sum_{i=0}^{N} (\tau^{(i+1)}-\tau^{(i)}) \delta_{\bt^{(i)}_{\bm{\psi}}(s,a)},
\end{equation}}  \par  \vspace{-.155in}
 \noindent{where} $\bt^{(i)}_{\bm{\psi}}$ is the quantile value at fraction ${\tau^{(i)}\! = \!\frac{i}{N}}$.
Dabney et al.~\cite{dabney2018distributional} showed that minimizing the 1-WD between the target ${T^{\pi} \! Z_{\overbar{\bm{\psi}}}(s,a) \! = \! r(\s,a)+ \! \gamma Z_{\overbar{\bm{\psi}}} (s', a')}$ and $Z_{\bm{\psi}}(s, a)$ can be achieved using quantile regression (QR) loss.  However, due to the QR loss non-smoothness, they minimized the quantile Huber loss with a threshold parameter $k  \!\!=\! \! 1$ for pairwise predicted quantiles $\bt^{(i)}_{\bm{\psi}}\!(s,a)$ and target quantiles $y^{(j)}\!(s,a) \!:= \! r(\s,a) \!+ \!\gamma \bt^{(j)}_{\overbar{\bm{\psi}}}(s',a')$:\footnote{\vspace{-.04in} Throughout the rest of the paper, for brevity, we omit the input notation $(s,a)$ when referring to $\bt^{(i)}_{\bm{\psi}}(s,a)$ and $y^{(j)}(s,a)$.}
\small{ \vspace{-.09in}{
\begin{equation}
\! \! \! \! \! \mathcal{L}^{\text{QR}}_{k=1}\!(\bm{\psi}) \! \! = \!\!  \frac{1}{N}\!\! \sum_{i=1}^N  \! \sum_{j=1}^N \! \left|\hat{\tau}^{(i)} \!- \!\delta_{\{ u^{(i,j)}<0 \}} \! \right| \frac{\mathcal{L}^{k=1}_{H}(u^{(i,j)})}{k},
\! \!   \label{Eq: w-distance} 
\end{equation}} }\normalsize \par  \vspace{-.1in}
 \noindent  where ${\hat{\tau}^{(i)} \!\!  = \!\! \frac{\tau^{(i-1)}\!+\tau^{(i)}}{2}}$, $u^{(i,j)} \!\!:= \!\! y^{(j)} \! -  \bt^{(i)}_{\bm{\psi}}$, and $\mathcal{L}^k_{H}\!(.)$  is the Huber loss with the threshold parameter $k$:
\vspace{-.12in}{
\begin{eqnarray}
\mathcal{L}^k_{H}(u)= \begin{cases}
    \frac{1}{2}u^2, & \text{if $|u|<k$}\\
    k(|u|- \frac{1}{2}k), & \text{otherwise}.
  \end{cases} \label{Eq:Huber}
\end{eqnarray}} \par  \vspace{-.1in}
\noindent Parameter $k$ influences the behaviour of the Huber (thus the quantile Huber) loss function in terms of being quadratic (smooth) near zero errors and linear (robust) for larger errors. It's important to note that although the Huber loss exhibits non-smoothness at $k$, this non-smoothness around $k$ is generally less severe than the non-smoothness of a purely linear loss around zero.
Moreover, while the Huber loss itself is symmetric, the term $|\hat{\tau}^{(i)} \! - \!\delta_{\{u^{(i,j)}<0\}}|$ in Eq.~\eqref{Eq: w-distance} introduces an asymmetry in the quantile Huber loss, enabling heterogeneous risk profile learning~\cite{zhao2021robust}. Hence, the quantile Huber loss is also called the asymmetric Huber loss.
\\
Several distributional RL methods~\cite{clavier2022robust,cao2023gamma, dabney2018implicit,  yang2019fully,  oh2023toward, he2022popo} have been developed using the quantile Huber loss with a fixed $k \!\!=\!\! 1$, neglecting the impact of varying $k$ on performance.
 \vspace{-.19in} \section{Generalized Quantile Huber Loss} \label{sec:loss}
 \vspace{-.11in}
 The WD is a symmetric metric measuring the minimum cost to transform one distribution into another~\cite{villani2009optimal}. The 1-WD between two single Dirac deltas, $\delta_{x_1}$ and $\delta_{x_2}$, using a cost function $\frac{\mathcal{L}^k_{H}(u)}{k}$, is given by $W_1(\delta_{x_1}\!,\!\delta_{x_2} ) \!\!= \!\! \frac{\mathcal{L}^k_{H}(|x_1-x_2|)}{k}$~\cite{villani2009optimal}.
Therefore, we can interpret the term $\frac{\mathcal{L}^{k}_{H}(u^{(i,j)})}{k}$ in Eq.~\eqref{Eq: w-distance} as the 1-WD between $p( {y^*}^{(j)}|y^{(j)})\!\!= \!\!\delta_{y^{(j)}}$ and $p({\bt^*}^{(i)}|\bt^{(i)}_{\bm{\psi}})\!\!= \!\! \delta_{\bt^{(i)}_{\bm{\psi}}}$, where ${y^*}^{(j)}$ and ${\bt^*}^{(i)}$ are the real quantiles of $T^{\pi}\!Z_{\overline{\bm{\psi}}}$ and $Z_{{\bm{\psi}}}$, respectively. Thus, $\mathcal{L}^{\text{QR}}_{k}$ in Eq.~\eqref{Eq: w-distance} can be expressed  as the average of asymmetric 1-WDs between these pairwise Dirac delta distributions:
\small{  \vspace{-.12in}{
\begin{eqnarray}
\lefteqn{ \!\!\! \!\!\!\!\!\! \!\!\!  \mathcal{L}^{\text{QR}}_{k}(\bm{\psi}) \!= \! \frac{1}{N}\!   \sum_{i=1}^N \!  \sum_{j=1}^N   \left|\hat{\tau}^{(i)}- \delta_{\{ u^{(i,j)}<0 \}} \right| \nonumber } \\
&&  W_1\! \left( \! p( {y^*}^{(j)}|y^{(j)}), p({\bt^*}^{(i)}|\bt^{(i)}_{\bm{\psi}}) \! \right). \label{Eq:sum_WD}
\end{eqnarray}} \par  } \normalsize
\vspace{-.11in}
\noindent
%
$p({\bt^*}^{(i)}|\bt^{(i)}_{\bm{\psi}})$  and $p( {y^*}^{(j)}|y^{(j)}\!)$ indeed model the noises (uncertainties) in the predicted and target quantiles, respectively. Thus, $p( {y^*}^{(j)}|y^{(j)})\!\!= \!\!\delta_{y^{(j)}}$ and $p({\bt^*}^{(i)}|\bt^{(i)}_{\bm{\psi}})\!\!= \!\! \delta_{\bt^{(i)}_{\bm{\psi}}}$ in Eq.~\eqref{Eq:sum_WD} show zero noises  in the quantile values.\\
However, as real-world data generally exhibits some level of noise and uncertainty, we assume the presence of independent zero-mean Gaussian noises $\epsilon_1$ and $\epsilon_2$, with standard deviations $\sigma_1$ and $\sigma_2$, affecting target and predicted quantiles. Specifically, we have ${y^*}^{(j)}\!\!= \!\!{y}^{(j)} \!+ \! \epsilon_1$ and ${\bt^*}^{(i)}\!\!= \!\!{\bt}^{(i)} \!+\! \epsilon_2$; hence, ${p({\bt^*}^{(i)}|\bt^{(i)}_{\bm{\psi}}\!) \! \! =\!\!  \mathcal{N}(\bt^{(i)}_{\bm{\psi}}\! ,\sigma_1 \! )}$ and  ${p({y^*}^{(j)}|y^{(j)}\!)\! \! =\!\!  \mathcal{N}(y^{(j)}\! ,\sigma_2 \! )}$. This allows us to compute $W_1$ as per~\cite{chhachhi20231}, as follows:
\small{ \vspace{-.09in}{ \begin{eqnarray}
\lefteqn{\!\!\! \!\!\!\!    W_1 \left(p({\bt^*}^{(i)}|\bt^{(i)}_{\bm{\psi}}), p({y^*}^{(j)}|y^{(j)}) \right)  } \nonumber \\
&& \!\!\! =  \left|\bt^{(i)}_{\bm{\psi}}-y^{(j)} \right| \! \left[1-2\phi_{N} \left(- \frac{|\bt^{(i)}_{\bm{\psi}}-y^{(j)}|}{|\sigma_1-\sigma_2|} \right) \! \right] \nonumber \\
 && \!\!\! +\left| \sigma_1-\sigma_2\right| \sqrt{\frac{2}{\pi}}\text{exp}  \!    \left(\! - \frac{(\bt^{(i)}_{\bm{\psi}}-y^{(j)})^2}{2 (\sigma_1-\sigma_2)^2} \! \right), \label{Eq:w_Gauss} 
\end{eqnarray}} } \normalsize \par \vspace{-.15in}
\noindent where $\phi_{N}$ is the cumulative distribution function (CDF) of the standard normal Gaussian distribution. 
To ensure a minimum value of zero at $\bt^{(i)}_{\bm{\psi}}-y^{(j)} \!\! = \!\! 0$, we add a constant term to Eq.~\eqref{Eq:w_Gauss} and propose the following cost function:
\small{ \vspace{-.1in}{  \begin{equation}
\!\!\!  C_{GL}^b \!(u) \!\! = \!\! |u|\! \bigg[1 \!\! - \!\!2\phi_{N}(- \frac{|u|}{b}) \! \bigg] \!\!+\! b \sqrt{\frac{2}{\pi}} \text{exp} \left(\!\! - \frac{u^2}{2 b^2} \right)\!\!-\! b \sqrt{\frac{2}{\pi}}, \!\! \nonumber
\end{equation}} } \normalsize \par  \vspace{-.185in}
\noindent where $b\!= \!|\sigma_1 \! - \! \sigma_2|$. Thus, $\bm{\psi}$ can be learned by minimizing the loss function $\mathcal{L}^{\text{GL}}_{b}(\bm{\psi})$ defined as:
\small{ \vspace{-.12in}{
\begin{eqnarray}
  \!\!    \mathcal{L}^{\text{GL}}_{b} (\bm{\psi}) \!= \!\! \frac{1}{N}\!   \sum_{i=1}^N   \sum_{j=1}^N   |\hat{\tau}^{(i)}- \delta_{\{ u^{(i,j)} <0 \}}| C_{GL}^b (u^{(i,j)}).  \!\!\!\!\! \label{Eq:final_loss}
\end{eqnarray}} } \normalsize \par 
\vspace{-.1in}
\noindent Parameter $b$ in $\mathcal{L}^{\text{GL}}_{b}(\bm{\psi})$ is the disparity between the standard deviations of predicted and target quantile noises. It is worth noting that although our loss function $\mathcal{L}^{\text{GL}}_{b}(\bm{\psi})$ is non-smooth near zero, it incorporates an exponential and CDF term, enhancing smoothness around zero compared to the Huber loss at its threshold.   Moreover, our loss function is less sensitive to outliers than Huber quantile loss due to the diminishing effect of the exponential term and the bounded penalties imposed by the CDF.
\vspace{.03in} \\
 \textbf{Relationship} \textbf{to the quantile Huber loss:} Using a second-order Taylor approximation for small $\frac{|u|}{b}$ and setting ${\text{exp} (\! - \frac{u^2}{2 b^2} )} \!\!= \!\! 0$ for large $\frac{|u|}{b}$, an approximation of $\mathcal{L}^{\text{GL}}_{b}(\bm{\psi})$, denoted as $\mathcal{L}^{\text{GL-A}}_{b}(\bm{\psi})$, is obtained:
\small{ \vspace{-.125in}{
\begin{eqnarray}
 \mathcal{L}^{\text{GL-A}}_{b}\! (\bm{\psi}) \! = \!\! \frac{1}{N}\!\!   \sum_{i=1}^N \!  \sum_{j=1}^N  \!  \left|\hat{\tau}^{(i)}- \delta_{\{ u^{(i,j)}<0 \}} \! \right|  \!  C_{GL-A}^b(u^{(i,j)}),\!\!\!\! \nonumber
\end{eqnarray}} } \normalsize \par  \vspace{-.16in}
\noindent
where 
\small{ \vspace{-.15in}{
\begin{eqnarray}
\!\!  C_{GL-A}^b(u)= \begin{cases}
    \frac{1}{b \sqrt{2 \pi}} u^2, & \text{if $|u|<b$}\\
    |u|-b \sqrt{\frac{2}{\pi}}, & \text{otherwise}.
  \end{cases} \label{Eq:Huber_A}
\end{eqnarray}} } \normalsize \par 
\vspace{-.17in}
\noindent 
Comparing Eq.~\eqref{Eq:Huber_A} with Eq.~\eqref{Eq:Huber}, we see $C_{GL-A}^b(u) \! \approx \! \frac{\mathcal{L}^k_{H}(u)}{k}$, implying  $ \mathcal{L}^{\text{GL-A}}_{b}(\bm{\psi}) \!\! \approx \! \!\mathcal{L}^{\text{QR}}_{k}(\bm{\psi}) $ for $k\!\!=\!\! b$. Hence, we can pick an appropriate $k$ value in $\mathcal{L}^{\text{QR}}_{k}$ as $k\!\!=\!\! b\!\!=\!\! |\sigma_1 \! - \! \sigma_2|$. As our proposed loss function $\mathcal{L}^{\text{GL}}_{b}$ includes the quantile Huber loss as its approximation, we refer to it as the generalized quantile Huber loss. 
 \vspace{-.2in} \section{Experiments} \label{sec:experiment}
\label{sec:illust}
 \vspace{-.15in} 
In this section, we compare our generalized loss function to the quantile Huber loss and explore how their relationship can enhance the latter's performance by aiding in threshold parameter selection.
\vspace{-.1in} \subsection{Atari Games} \vspace{-.1in}
While keeping all other components unchanged, we replace $ \mathcal{L}^{\text{QR}}_{k=1}\! (\bm{\psi})$ in QR-DQN~\cite{dabney2018distributional} and FQF~\cite{yang2019fully} with $\mathcal{L}^{\text{GL}}_b(u)$ and $\mathcal{L}^{\text{GL-A}}_b(u)$, creating new algorithms: GL-DQN, GLA-DQN, GL-FQF, and GLA-FQF.
 We assess these approaches across $55$ Atari games, as in QR-DQN and FQF.  We estimate $\sigma_1$ and $\sigma_2$ by calculating the sample standard deviation averaged across all batches and set $b \!\!=\!\!|\sigma_1-\sigma_2|$.
\\
Table~\ref{tab:atari_results} compares mean and median human-normalized scores used in Atari game RL literature.  GLA-DQN outperforms QR-DQN, and GLA-FQF performs comparably to FQF, showing the effectiveness of our $k$ interpretation.  Moreover, GL-DQN and GL-FQF surpass  GLA-DQN and GLA-FQF, respectively, showing faster convergence due to the added smoothness and robustness in the generalized loss function $\mathcal{L}^{\text{GL}}_b(u)$ due to the exponential and CDF terms.
\vspace{-.15in} 
\subsection{Gamma and Vega Hedging Results}
\begin{figure}[t]
    \centering
    \includegraphics[scale=0.355]{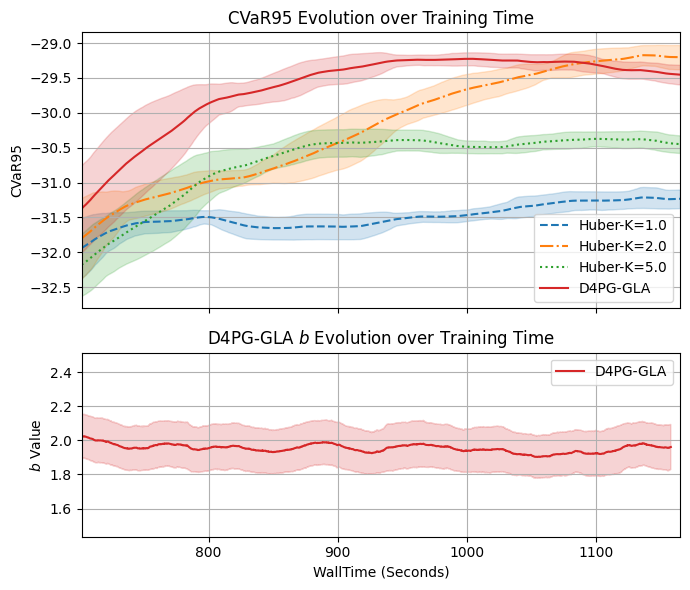}
  \vspace{-.12in}   \caption{Hedged portfolio's CVaR95 and D4PG-GLA $b$-value evolution during training.}
    \label{fig:results} 
\end{figure}

\vspace{-.1in} 
To test our interpretation for $k$ in a more realistic setting, we assess its performance in option hedging, a risk-aware financial application where the objective is to optimize the Conditional Value-at-Risk at a $95\%$ confidence level (CVaR95) for total rewards. We focus on the SABR model with a $0.5\%$ transaction cost in a market simulated based on the SABR stochastic model~\cite{sabr}, as described in the work by Cao et al.~\cite{cao2023gamma}, Section~4.4.
 We replace $\mathcal{L}^{\text{QR}}_{k}$ in D4PG-QR~\cite{cao2023gamma} with $\mathcal{L}_b^{\text{GL-A}}$ using the estimated $b \!\!= \!\!|\sigma_1 \!- \!\sigma_2|$, creating D4PG-GLA. \footnote{The code is available \href{https://github.com/pmalekzadeh/A-robust-quantile-huber-loss}{here}. }
 \\
 Fig.~\ref{fig:results} depicts the training curves, indicating that D4PG-GLA converges faster than D4PG-QR with various $k$ values.  Interestingly, the optimal value of $k$ for D4PG is not  $1$, and D4PG-GLA's estimated $b$-value during training aligns with this optimal $k$ ($\approx \!\! 2$). These results emphasize the advantages of fine-tuning $k$ in the quantile Huber loss and demonstrate the effectiveness of our $k$ interpretation, reducing the need for extensive parameter searches.  This approach is valuable in option hedging, where market misspecification is already a significant challenge. D4PG-GLA experiences a performance decline in the final stages of training, but using a stopping criterion allows us to finish training earlier than D4PG-QR while achieving similar or better CVaR levels.
\\
\vspace{-.18in} \begin{table} [t] 
\centering
 \caption{Learning scores for $55$ Atari games.} \vspace{-.15in}
\begin{tabular}{|c c c c|} 
 \hline
 Method & Mean & Median & $>$ Human \\ [0.5ex] 
 \hline
 QR-DQN & 902\% & 193\% & 41 \\ 
 \hline
 GL-DQN & \textbf{934\%} & \textbf{209\%} & \textbf{42} \\
 \hline
 GLA-DQN & 917\% & 204\%  & 41 \\
 \hline
  FQF & 1426\% & 272\% & {44} \\ 
 \hline
 GL-FQF & \textbf{1443\%} & \textbf{281\%} & \textbf{44} \\
 \hline
 GLA-FQF & 1435\% & 275\% & {44} \\ 
 \hline
\end{tabular}
    \label{tab:atari_results}
    \vspace{.02in} 
\end{table}
\vspace{-.0in}    \vspace{.04in}
\vspace{-.15in}  \section{Conclusion} \vspace{-.1in} \label{sec:conclusion}
We provide a probabilistic interpretation of the quantile Huber loss, resulting in a more robust generalized quantile Huber loss that includes the traditional version and facilitates easy threshold parameter adjustments.
Recent distributional RL work~\cite{cao2023gamma, clavier2022robust, oh2023toward, he2022popo} uses the quantile Huber loss with a fixed threshold parameter $k\! \!=\! \!1$. Our approach enables parameter adaptation in RL tasks without exhaustive grid searches.
\vfill
\pagebreak
\bibliographystyle{IEEEbib}
{\footnotesize \bibliography{refs}}

\end{document}